\documentclass[runningheads]{llncs}
\usepackage{graphicx}
\usepackage{amsmath,amssymb} 
\usepackage{color}
\usepackage{url}
\usepackage[width=122mm,left=12mm,paperwidth=146mm,height=193mm,top=12mm,paperheight=217mm]{geometry}

\begin{document}
\title{Image Generation from Sketch Constraint Using Contextual GAN} 

\titlerunning{Image Generation from Sketch Constraint Using Contextual GAN}
%

\author{Yongyi Lu\inst{1}\thanks{This work was partially done when Yongyi Lu was an intern at Tencent Youtu.} \and
Shangzhe Wu\inst{1,3} \and
Yu-Wing Tai\inst{2} \and
Chi-Keung Tang\inst{1}}


%
\authorrunning{Y. Lu, S. Wu, Y.W. Tai and C.K. Tang}
%


\institute{The Hong Kong University of Science and Technology \and
Tencent Youtu \and
University of Oxford\\
\email{\{yluaw,cktang\}@cse.ust.hk, swuai@connect.ust.hk, yuwingtai@tencent.com}}

\maketitle              
\begin{abstract}
In this paper we investigate image generation guided by hand sketch. 
When the input sketch is badly drawn, the output of common {\em image-to-image translation} 
follows the input edges due to the hard condition imposed by the translation process. 
Instead, we propose to use sketch as weak constraint, 
where the output edges do not necessarily follow the input edges.  
We address this problem using a novel {\em joint image completion} 
approach, where the sketch provides the image context for completing,
or generating the output image. We train a generated adversarial network, i.e, 
{\em contextual GAN} to learn the joint 
distribution of sketch and the corresponding image by using joint images. 
Our contextual GAN has several advantages. 
First, the simple joint image representation allows for simple and effective learning of 
joint distribution in the same image-sketch space, which avoids complicated issues in 
cross-domain learning. Second, while the output is related to its input overall, 
the generated features exhibit more freedom in appearance and do 
not strictly align with the input features as previous conditional GANs do. 
Third, from the joint image's point of view, 
image and sketch are of no difference, thus exactly the same deep joint image completion 
network can be used for image-to-sketch generation. Experiments evaluated on 
three different datasets show that our contextual GAN can generate more realistic images than 
state-of-the-art conditional GANs on challenging inputs and generalize well on common categories.


\end{abstract}
\section{Introduction}

\begin{figure}[t]
	\begin{center}
		\includegraphics[width=0.8\linewidth]{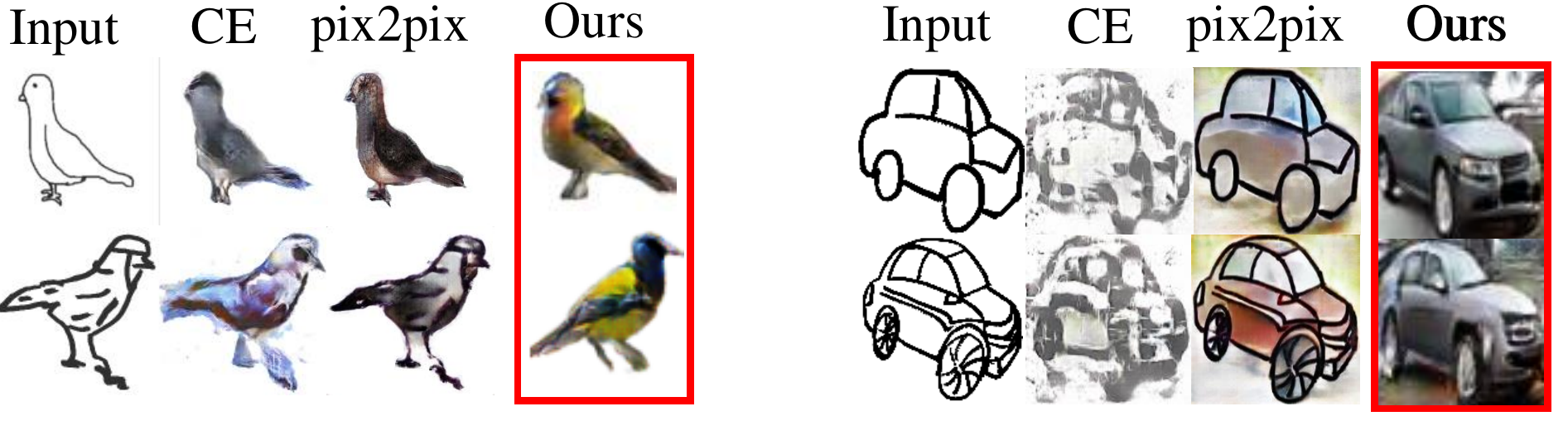} \\
	\end{center}
	\caption{Freehand sketch to image results by two conditional GANs (i.e., CE~\cite{contextencoder}, and pix2pix~\cite{isola2016image}) and our contextual GAN. Even the sketches are badly drawn they are still expressive in conveying features of birds and cars. Our method does not require strict alignment while still faithful to the input, resulting in more realistic images.} 
	\label{fig:teaser}
\end{figure}

Image translation generates impressive photographic results in a variety of applications
demonstrated in~\cite{isola2016image}. Common approaches of conditional generated adversarial networks (cGAN) incorporate hard condition like pixel-wise correspondence~\cite{isola2016image} alongside the translation process, which makes the output strictly align with the input edges.  This can be highly problematic in sketch-to-image generation when the input is a free-hand sketch. Figure~\ref{fig:teaser} shows casual 
sketches of different objects, where the free-hand sketch and its natural photographic object
do not strictly align to each other. In such case, conditional GAN (e.g., pix2pix~\cite{isola2016image}) is incapable of generating realistic and visually comfortable images. 

Our goal of sketch-to-image generation is to automatically generate a photographic image 
of the hand-sketched object. Even a poorly drawn sketch allows non-artists to easily specify an object's attributes in many situations which may be clumsy to specify in verbose text description. On the other hand, the translation should respect the sparse input content, but might need some deviation in shape to generate a realistic image.

\begin{figure}[t]
	\begin{center}
		\includegraphics[width=0.9\linewidth]{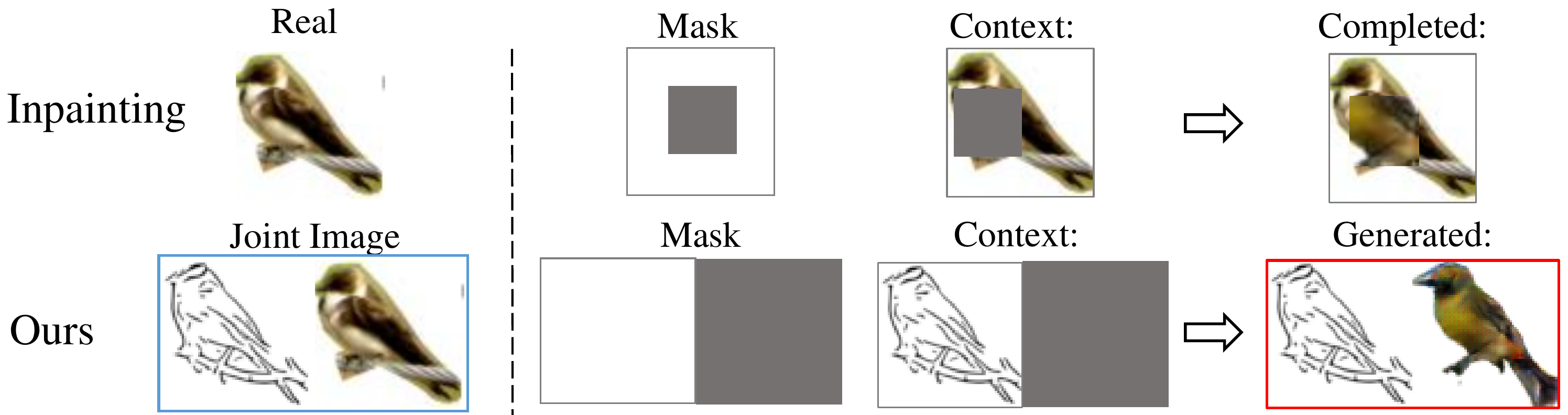} \\
	\end{center}
	\caption{Image generation posed as image completion. Top:~Semantic image inpainting with the uncropped part as context~\cite{deep_completion}. Bottom:~Joint image completion with the sketch as the ``uncropped" context, with the cropped part on the right. Our joint image concatenates a sketch and its corresponding image side-by-side.} 
	\label{fig:overview}
\end{figure}

\begin{figure}[t]
	\begin{center}
		\includegraphics[width=1.0\linewidth]{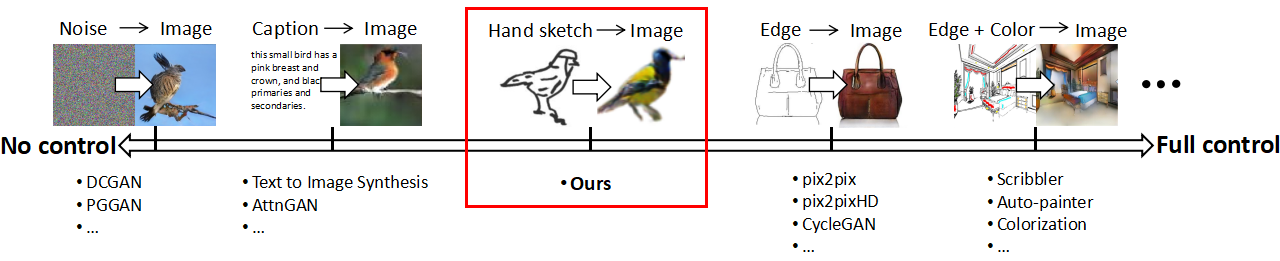} \\
	\end{center}
	\caption{Illustration of the whole GAN spectrum in image generation task. Our contextual GAN is in stark contrast to unconditional GAN and conditional GAN in that we use sketch as context (a weak constraint) instead of generating from noise or with hard condition, which has not been well studied in previous approaches.} 
	\label{fig:spectrum}
\end{figure}

In order to tackle these challenges, we propose a novel contextual generative adversarial network for sketch-to-image generation.
We pose the image generation problem as an image {\em completion} problem, with sketch providing a weak contextual constraint. Figure~\ref{fig:overview} illustrates the core concept.
In conventional image completion, the corrupted part of an input image is 
completed using surrounding image content as context.  In the bird completion example, the unmasked, 
partial bird features are the contextual information.  By analogy,
in our joint sketch-image completion, the ``corrupted" part consists of the 
entire image to be generated, while the ``context" is provided by the input sketch 
for the ``completion" or generation of the photographic object. In this way, we change our objective from common image-to-image translation in conditional GAN (sketch as hard condition) to completing the missing entire image in joint image completion (sketch as context). Please refer to Figure~\ref{fig:spectrum} to get a sense of the difference between our proposed contextual GAN and common conditional GANs in the whole GAN spectrum.

Based on this novel joint sketch-image completion idea we propose a new and simple contextual GAN framework. A generative adversarial network is trained to learn the joint distribution and capture the inherent correspondence between a sketch and its corresponding image using the defined joint image. This approach encodes the ``corrupted" joint image into the closest ``uncorrupted" joint image in the latent space via back propagation, which can be used to predict and hence generate the output image part of the joint image. To infer a closest mapping, we use sketch as a weak constraint and define our objective function which is composed of a contextual loss as well as traditional GAN loss. We also propose a straight-forward scheme for better initialization of sketch. 


This novel approach has several advantages: (1) there are
no separate domains for image and sketch learning; only one 
network is used to understand  a joint sketch-image pair
which is a single image. This is in stark contrast with image translation where only sketch is treated as input.  
(2) By using weak sketch constraint, while related to its input edges, the generated image 
may exhibit different poses and shapes beyond the input sketches which
may not strictly correspond to photographic objects. 
(3) From the joint image's point of view, image and sketch are of no difference, so
they can be swapped to provide the context for 
completion for the other.  Thus, exactly the same sketch-to-image generation 
approach/network can be used for the reverse or image-to-sketch generation.

Our proposed framework is generic which can employ any state-of-the-art 
generative model. Capitalizing on the GAN for image completion~\cite{deep_completion}, 
we propose a two-phase recipe to learn the sketch-image 
correspondence inherent in a joint image as well as imposing the weak sketch constraint. For training, the network learns the sketch-image correspondence using uncropped joint images. In completion, we search for an encoding of the provided corrupted image 
using only the sketch to provide the weak context for ``completing" 
the image based on a modified objective. This encoding is then used to reconstruct the image by feeding it to the generator which generates the photographic object from sketch. Experimental results show that our contextual GAN can generate more realistic and natural images than state-of-the-art conditional GANs on challenging inputs, e.g. poorly drawn sketches shown in Figure~\ref{fig:teaser}, while producing comparable results to state-of-the-arts on good sketches where the edges correspond to photographic objects.

\section{Related Work}
The rapid development of deep learning has accounted for 
recent exciting progress in image generation,
especially the introduction of generative 
adversarial networks (GAN)~\cite{GAN}. 
Conditioning variables were then introduced to GAN~\cite{pixelCNN,SSGAN,stackGAN}. 
Related to our contextual GAN for joint images 
is deep image completion with perceptual and contextual 
losses~\cite{deep_completion}. Pretrained with uncorrupted data, 
the $G$ and $D$ networks are trained to reconstruct a complete image.  
Their 
impressive examples show that even a large region of a facial image 
is cropped from the input, 
the generated complete facial image looks very realistic. 
Another impressive work on image completion~\cite{contextencoder}
is based on autoencoders with a standard reconstruction loss
and an adversarial loss.   
Autoencoders have also been successfully applied to generating images from
visual attributes~\cite{attribute2image}.
Analogous to image completion, where the uncropped part of the 
image provides the proper context for facial image 
completion, in our sketch-to-image 
generation, the {\em entire} input sketch is regarded as the ``uncropped context"
for completing the entire natural image part of the joint image.  

Another way of generating images from sketches requires a huge database from which images are retrieved. In~\cite{sketchmeshoe}, a database 
of sketch-photo pairs was collected, and deep learning was used to 
learn a joint embedding. The Sketchy Database~\cite{sketchyDB} contributed 
a collection of sketch-photo pairs which were used to train a cross-domain CNN
to embed them in the same space.  
In recent works, the mapping between sketches and images 
was studied in sketch-based image retrieval, where the sketch and
the image were learned in separate networks. In~\cite{triplet}, 
several triplet CNNs were evaluated for measuring 
the similarity between sketches and photos. Triplet networks 
are used to learn joint embeddings. 
While classical representations were proposed to retrieve images 
from sketch queries~\cite{kato1992,delBimboP97,Sclaroff97},
recent methods used sophisticated feature 
representations~\cite{CaoWZZ11,shrivastava-sa11,Cao_2013_ICCV,BMVC2015_164}. 
Recent cross-domain embedding methods trained deep networks to learn 
a common feature space for sketches and 3D models~\cite{WangKL15},
and images and 3D models~\cite{li2015}. Siamese networks trained 
with contrastive loss~\cite{chopra2005} and triplet or ranking 
loss~\cite{wang2014} were proposed.  On the other hand,
we do not require such sketch-photo collections or 
regard sketch and image as two separate domains,
as they form the same joint image.  This allows for an 
effective encoding for the ``corrupted" joint image in 
the latent space  and leads to stable training.

\section{Approach}
\label{sec: approach}

Recent work in semantic inpainting regards inpainting as a constrained image generation problem~\cite{deep_completion}, where the generated content is supposed to be well aligned with the surrounding pixels while maintaining semantic reality based on the observed context. Analogously, we pose the image generation problem as an image {\em completion} problem, with sketch providing a weak contextual constraint. Our deep model is built on the GAN architecture proposed 
in~\cite{DBLP:journals/corr/RadfordMC15} with the following technical revisions.  

\subsection{Joint Sketch-Image Representation}
\label{sec:joint}
Sketch-to-image generation is a nontrivial task since sketches are often highly abstract with sparse 
visual content, and they are sometimes badly drawn. Rather than following traditional ways of separating sketch and image, we propose to model them in a joint input space. Specifically, based on a 
corpus containing samples with real images (\textbf{B}) and their aligned sketch styles (\textbf{A}), we spatially concatenate them into a joint sketch-image pair (\textbf{AB}), as shown in Figure~\ref{fig:overview}. In our framework, the joint image naturally captures the contextual information, i.e., correspondence between the sketch and image portions, which is effective for learning their joint distribution using GAN. 

\begin{figure}[t]
	\begin{center}
		\begin{tabular}{cc}
			\includegraphics[height=3.5cm]{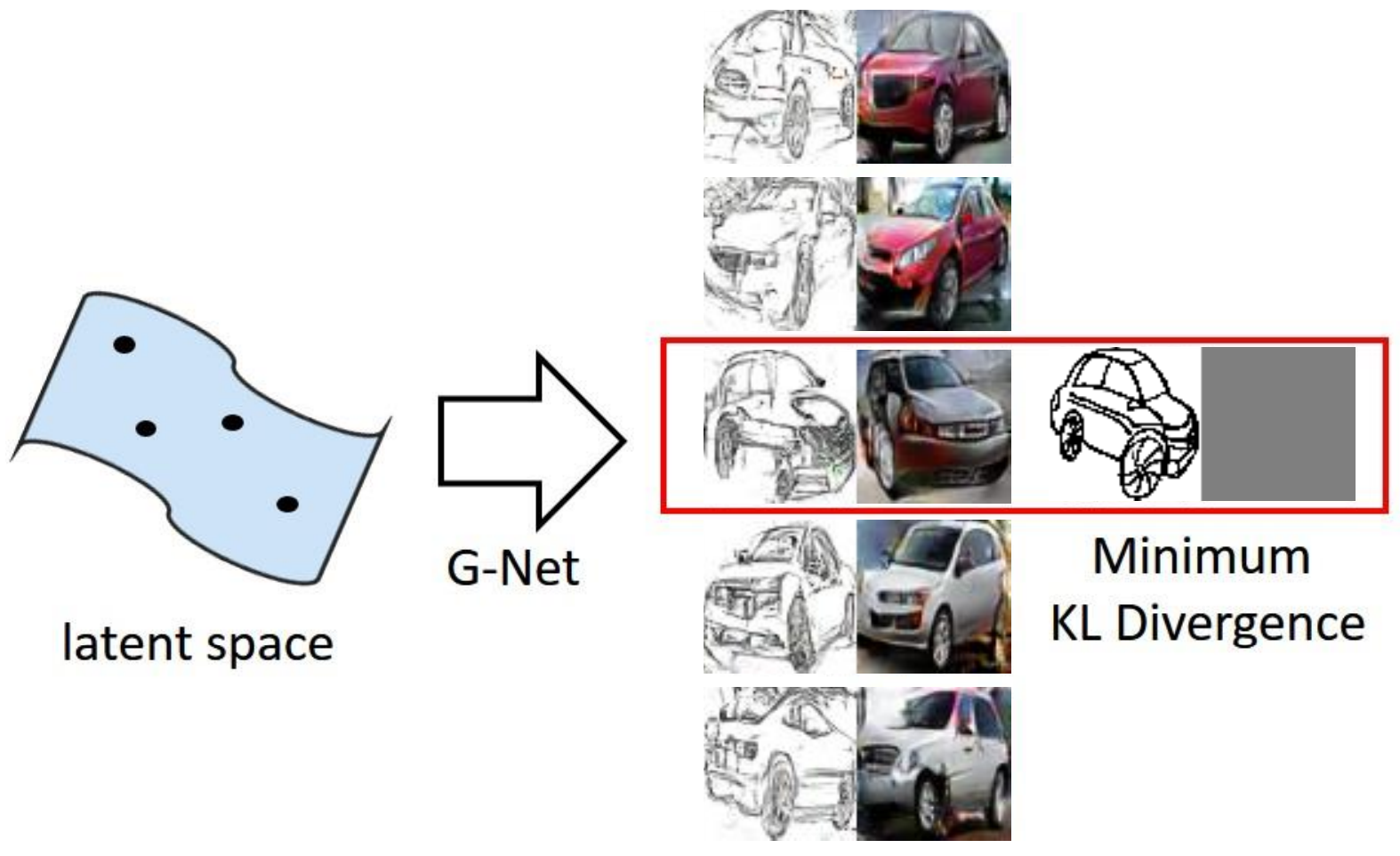} &
			\includegraphics[height=3.5cm]{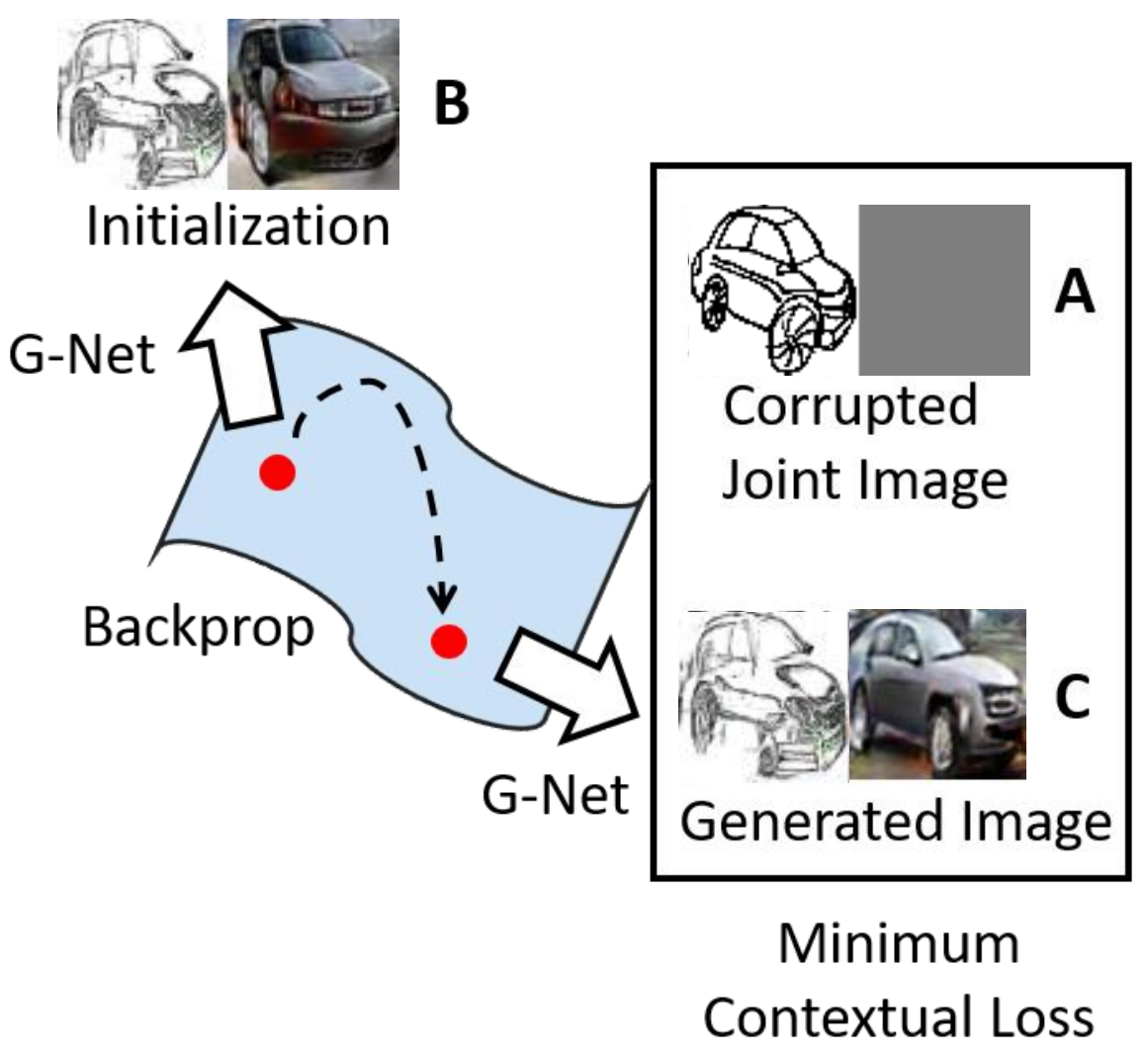}  \\
			(a) Refined initialization.   &  (b) Completion pipeline.
		\end{tabular}
	\end{center}
	\caption{Contextual GAN pipeline. (a)~Initialization: red box represents better initialization based on KL-divergence. (b) Given the initialization $B$, we use back propagation on the loss defined in Eq.~(\ref{eq:total_loss}) to map the corrupted image $A$ to the latent space. The mapped vector is then passed through the $G$ network to generate the missing image $C$.}
	\label{fig:stage2}
\end{figure}

Specifically, we train a GAN model using joint images, the generator then automatically predicts the corrupted image part based on the context of the corresponding sketch part. The generator embeds the joint images onto a non-linear joint space $z$, i.e., $z$ is a joint embedding of sketch and image, whereas in previous work (e.g., ~\cite{zhu2016generative}) $z$ is only an embedding of image. As a result, instead of constraining the generated image directly with entire $z$ (hard constraint), we are able to constrain the generated image indirectly via the sketch part of the joint embedding $z$ of the input, thus remains faithful while exhibiting some degree of freedom in the appearance of the output image. Figure~\ref{fig:stage2} illustrates this pipeline which will be detailed in subsequent sections.

\subsection{Objective Function}
\label{sec:dcc}

To get the closest mapping of the corrupted joint image and the reconstructed joint image, we need to search for a generated joint image $G(\hat{z})$ in which the sketch portion best resembles the input sketch. Given the randomly sampled input $z \sim p_z$, we define our loss function to comprise of two losses in our objective:\\

\noindent {\bf Contextual Loss. } We use a contextual loss~\cite{DBLP:journals/corr/RadfordMC15} to measure the context similarity between the uncorrupted portions, i.e., the input sketch portion and the reconstructed sketch, which is defined as:
\begin{equation}
\label{eq:contextual}
\begin{split}
\mathcal{L}_{\mathit{contextual}}(z) = D_{KL}(\textbf{M} \odot y, \textbf{M}\odot G(z))
\end{split}
\end{equation}
where $\textbf{M}$ is the binary mask of the corrupted joint image and $\odot$ denotes the Hadamard production. Different from~\cite{DBLP:journals/corr/RadfordMC15}, since a sketch is a binary image rather than a natural image, we use the KL-divergence to measure the similarity between the distribution of two sketches which tends to produce better alignment of sketches. Ideally, all the pixels at the sketch portions are the same between $y$ and $G(z)$, i.e., $\mathcal{L}_{\mathit{contextual}}(z) = 0$, and we penalize $G(z)$ for not generating a sketch similar to the observed input sketch $y$.\\

\noindent{\bf Perceptual Loss.} The perceptual loss maintains the semantic content of the predicted image, which is defined using the adversarial loss of the $G$ network:
\begin{equation}
\label{eq:perceptual}
\begin{split}
\mathcal{L}_{\mathit{perceptual}}(z) = \log (1-D(G(z)))
\end{split}
\end{equation}

The objective function for $\hat{z}$ is then the weighted sum of the two losses:
\begin{equation}
\label{eq:total_loss}
\begin{split}
\hat{z} = \arg \min \limits_{z}(\mathcal{L}_{\mathit{contextual}}(z) + \lambda \mathcal{L}_{\mathit{perceptual}}(z))
\end{split}
\end{equation}
where $\lambda$ is a hyperparameter to constrain the generated image with the input. A small $\lambda$ will guarantee similar appearance of the input and output.

\subsection{Contextual GAN}
\label{sec:two-stage}
Our contextual GAN consists of the training stage and completion stage. The training stage is the same as the traditional GAN training except that our training samples are joint images. After training, we learn a generative network $G$ that achieves the objective of reproducing the joint image data distribution, i.e., mapping samples from noise distribution $p_z$ to the data distribution $p_{data}$. \\

\begin{figure}[t]
	\begin{center}
		\includegraphics[width=0.99\linewidth]{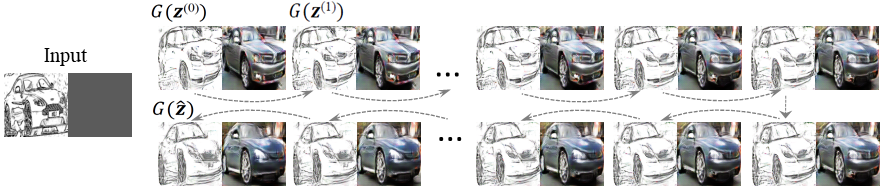} \\
	\end{center}
	\caption{Manifold traversing (with four iterations as shown) when updating latent vector $z$ using back-propagation. $\textbf{z}^{(0)}$ is random noise picked via our initialization scheme; $\textbf{z}^{(k)}$ denotes the $k$-th iteration; and $\hat{\textbf{z}}$ the final solution.} 
	\label{fig:manifold}
\end{figure}

\noindent {\bf Projection through Back Propagation. } Our goal is to encode the corrupted joint image input (i.e., the image portion that has been masked out) to the closest image on the manifold of $G$ in the latent space, so that we can use this closest joint image as our predicted results. Instead of maximizing $D(y)$, we compute the $\hat{z}$ vector that minimizes our objective function in Eq.~(\ref{eq:total_loss}). This means that we are projecting the corrupted input onto the $z$ space of the generator through the iterative back propagation. Specifically, the input is a vector $z$ initialized with uniformly random noise, and a joint image with only the sketch on the left with the image on the right being masked out. We back propagate the loss in Eq.~(\ref{eq:total_loss}) to update the randomly sampled input $z$ of network $G$. Note that in this stage only the input vector $z$ is updated using gradient descent, the weights of the network $G$ and $D$ remain unchanged. 
Figure~\ref{fig:manifold} provides visualization of traversing the latent space during back-propagation (with four iterations as shown). Note that~\cite{deep_completion} also adopts similar gradient descent optimization on inverse mapping.

After back-propagation, the corrupted input $y'$s closest mapping vector $\hat{z}$ in the latent space is then passed through $G$ network to generate $G(\hat{z})$. The resulting image fills in the missing values of $y$ (the image portion) with $G(\hat{z})$:
\begin{equation}
\label{eq:predict}
\begin{split}
\textbf{x}_{generated} = \textbf{M} \odot y + (1-\textbf{M}) \odot G(\hat{z})
\end{split}
\end{equation}

\noindent {\bf Initialization.} We use a uniformly sampled noise vector as input. An obvious problem is that the generated image is affected by the initialization. If the initialized sketch portion of $G(z)$ perceptually exhibits a large gap from the input sketch, it will be hard for the corrupted image to be mapped to the closest $z$ in the latent space with gradient descent. This will result in failure samples even if we set a very small $\lambda$ in Eq.~(\ref{eq:total_loss}). To address this problem, we refine the initialization as follows: we sample $N$ uniformly random noise vectors, and obtain their respective initialized sketches via the forward pass. Then we compute the pairwise KL-divergence between the input sketch and these $N$ initialized sketches. The one which gives the lowest KL-divergence represents the best initialization among the $N$ samples and will be used as the initialized sketch. 
This process is illustrated in Figure~\ref{fig:stage2}. We set $N=10$ in our implementation.

\subsubsection{Network Architecture}
\label{sec:network}
\begin{figure}[t]
	\begin{center}
		\includegraphics[width=0.6\linewidth]{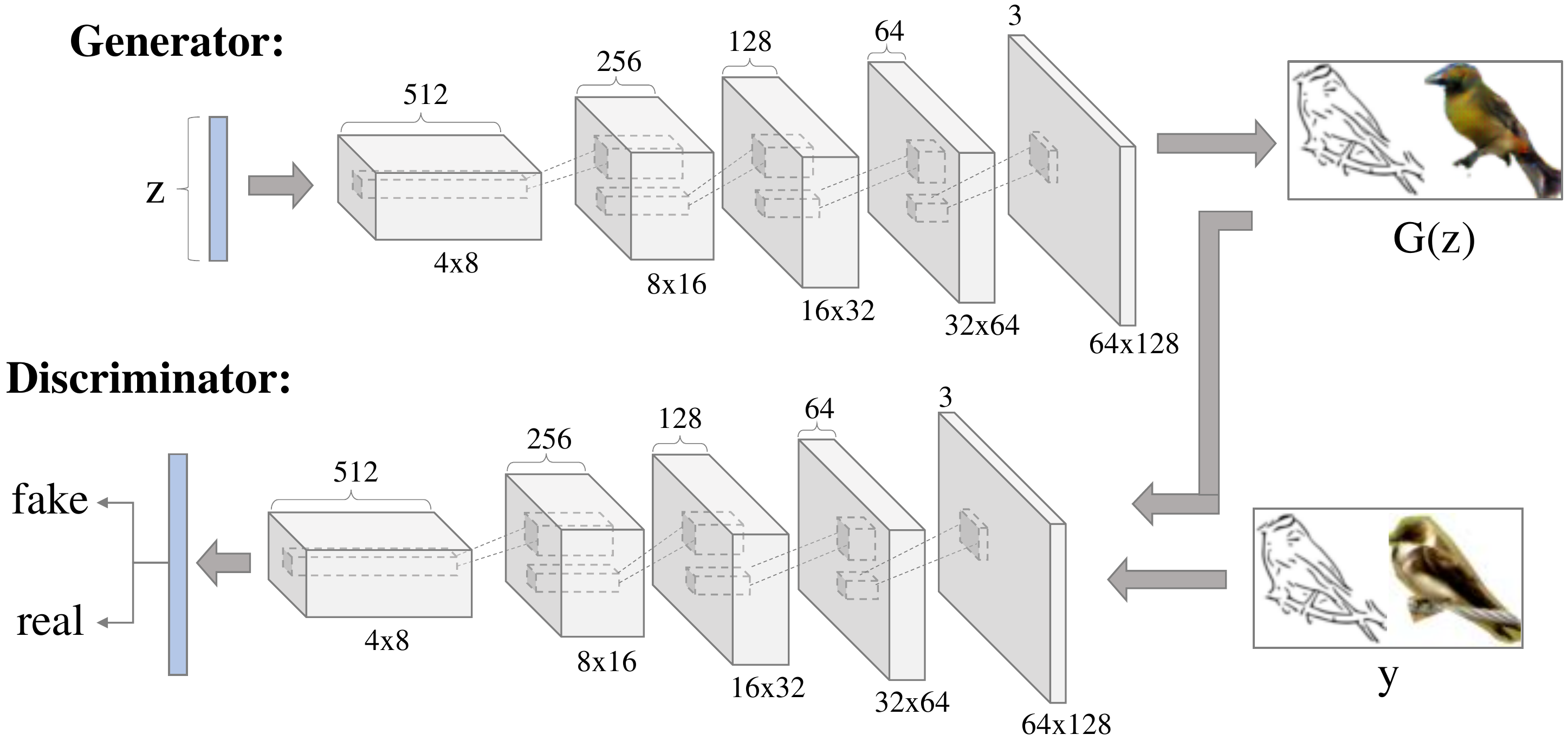}
	\end{center}
	\caption{The $G$ and $D$ network architecture for Contextual GAN.}
	\label{fig:framework}
\end{figure}

Figure~\ref{fig:framework} shows the complete network. Following~\cite{DBLP:journals/corr/RadfordMC15}, a $100$-D random noise vector, uniformly sampled from $-1$ to $1$, is fed into the generator $G$. Then, a $8192 \times 2$ linear layer reshapes the input to $4 \times 8 \times 512$. We use five up-convolutional layers with kernel size $5$ and stride $2$. We also have a batch normalization layer after each up-convolutional layer except the last one to accelerate the training and stabilize the learning. The leaky rectified linear unit (LReLU) activation is used in all layers. Finally,  $\tanh$ is applied in the output layer. This series of up-convolutions and non-linearities conduct a non-linear weighted upsampling of the latent space, and generates a higher resolution image of $64 \times 128$.

For the discriminator, the input is an image of dimension $64 \times 128 \times 3$, followed by 4 convolutional layers where the feature map's dimension is halved, and the number of channels is doubled from the previous layer. Specifically, we add 4 convolutional layers with kernel size $5$ and stride $2$ to produce a $4 \times 8 \times 512$ output. We then add a fully connected layer to reshape the output to one dimension, followed by a softmax layer for computing loss. 

\subsection{Network Generalization}
Real freehand sketches exhibit a large variety of styles and it may be very different from synthesis sketches automatically generated from images. To improve the network generality and to avoid overfitting to some particular style of sketch-image pairs, we augment our training data by using multiple styles of sketches as the training set. Specifically, we use the XDoG edge detector proposed in~\cite{Winnemller2012740}, the Photocopy effect~\cite{photocopy} in Photoshop and the FDoG filter proposed in~\cite{Kang:2007:CLD} to produce different styles of sketches. To better resemble hand-drawn sketches, we also simplify the edge images using~\cite{SimoSerraSIGGRAPH2016}. 

We split the data in each style into a training set and a testing set and train different style models. Instead of training all style models from scratch, we first obtain the pre-trained XDoG style model. The networks are then finetuned using sketches of other styles, i.e., the photocopy style, the simplification and FDoG. The reason is that we find XDoG is more similar to the original photographic image and contains more details. In this way, we can guarantee the network is first trained with a good local minima before augmenting the network with other sketch styles. We show in the experimental session that the augmenting styles help generalize the sketch-image correspondence better, while allowing some degree of freedom in appearance.

\section{Datasets and Implementation}\label{sec:exp}
In this section we describe the datasets used in training and implementations.

We tested our network using $3$ categories of images: face, bird, and car. Since the available sketch datasets are very limited, we applied several tools to produce sketches from images for training. We obtained the raw face, bird and car images from the \textit{Large-scale CelebFaces Attributes (CelebA) dataset}~\cite{Liu_2015_ICCV}, the \textit{Caltech-UCSD Birds-200-2011}~\cite{WahCUB_200_2011} dataset and the Stanford's \textit{Cars} Dataset~\cite{car_dataset}. 

\subsection{Data Preprocessing}
For the face category, the \textit{CelebA} dataset contains around $200$K images. We cropped and aligned the face region using OpenFace dlib~\cite{amos2016openface}. We obtained $400$K images with 2 different landmark maps for alignment. We generate three styles of sketches respectively and finally, we obtained $1200$K face sketch-image pairs. 


For the bird category, the \textit{CUB-200-2011} dataset contains only $11.7$K raw images. We first produced three styles of sketches using the above methods. To remove background as much as possible, we cropped the object and corresponding sketches based on the annotated bounding boxes. To augment this dataset for training, we randomly cropped $4$ images per image, and flipped them horizontally. Finally, we obtained around $100$K bird sketch-image pairs. 

For the car category, we simply used the $16$K car images from Stanford's \textit{Cars} Dataset~\cite{car_dataset} and produced one style of sketch. 


\subsection{Implementation}
\label{sec:implementation}
We pretrain the network for each category using contextual GAN. We use the Adam optimizer~\cite{adam} with a learning rate of $0.0002$ and a beta of $0.5$ for both the generator and discriminator network. The network is trained with a batch size of $64$ and epochs of $200$, which takes 6 to 48 hours for training depending on the size of the training set. After obtaining a well-trained model of XDoG style, we then finetune it using other styles of sketches sequentially at a lower learning rate (e.g., $1e^{-5}$) using the same network structure to obtain other styles' models.

The input $z$ is updated during completion, using a contextual loss and a perceptual loss with a $\lambda$ of $0.01$ and a momentum of $0.9$. Stochastic clipping is applied during back-propagation. We set a relatively small $\lambda$ so contextual loss is more important in test-time optimization so that the sketch portion in generated image best resembles the input sketch. The generator and the discriminator are fixed during back-propagation. For the experimental results, this update can be done in 500 iterations (the loss Eq.~(\ref{eq:total_loss}) converges very fast with our refined initialization, typically becomes stable after 100 iterations, which takes $<$1s). We use the same network architecture for all of the three categories. 

\section{Results}
In this section we will present our experimental results and comparison on the benchmark datasets described above. We also test our contextual GAN using in our opinion quite ugly hand-drawn sketches. These hand-drawn sketches are never presented in the training examples. 

\subsection{Benchmark Datasets}
As stated above, we first train a base model using the network we described above on one style, i.e., the XDOG style, and then finetune it on other styles: the photocopy effect and simplification style. 
This strategy is more effective compared to training with all different styles together. 
Ideally we can generate images using arbitrary styles of sketch providing that the pretrained model learns semantically correct content as well as faithful correspondence. We will test using hand-drawn sketches never seen by the network. 

\subsubsection{CelebA}
\begin{figure}[t]
	\begin{center}
		\footnotesize
		\begin{tabular}{ccccc}
			XDoG &
			\includegraphics[height=1.1cm]{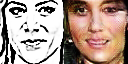} &
			\includegraphics[height=1.1cm]{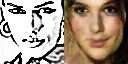} &
			\includegraphics[height=1.1cm]{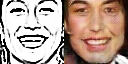} &
			\includegraphics[height=1.1cm]{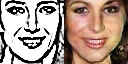} \\ 	
			PC & 
			\includegraphics[height=1.1cm]{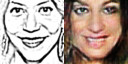} &
			\includegraphics[height=1.1cm]{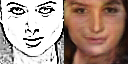} &
			\includegraphics[height=1.1cm]{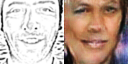} &
			\includegraphics[height=1.1cm]{figures/picked_results/face/ps/0675_0960.png} \\
			Sim &
			\includegraphics[height=1.1cm]{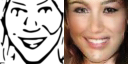} &
			\includegraphics[height=1.1cm]{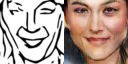} &
			\includegraphics[height=1.1cm]{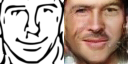} & 
			\includegraphics[height=1.1cm]{figures/picked_results/face/sim/0650.png} \\
		\end{tabular}
	\end{center}
	\caption{Results on CelebA dataset in three sketch styles: XDoG, Photocopy (PC) and simplified (Sim). Best viewed in color.}
	\label{fig:face_res}
\end{figure}

Figure~\ref{fig:face_res} shows some of our results on the CelebA dataset with three different styles, which demonstrates that our method can successfully predict or ``complete" the missing image part given the uncorrupted content, and generate a high-quality image that corresponds well to the given sketch. Note that the generated results not only capture the important details from and thus remains faithful to the input sketch, but also exhibit some degree of freedom in the appearance,
in comparison to state-of-the-art image generation methods such as pixel-to-pixel approach~\cite{isola2016image} where the generated results conform strictly to the input sketch's edge profile. 



\subsubsection{CUB}
\begin{figure}[!t]
	\begin{center}
		\footnotesize
		\begin{tabular}{ccccc}
			XDoG &
			\includegraphics[height=1.1cm]{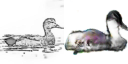} &
			\includegraphics[height=1.1cm]{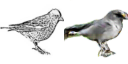} &
			\includegraphics[height=1.1cm]{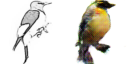} &
			\includegraphics[height=1.1cm]{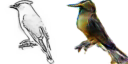} \\
			PC & 
			\includegraphics[height=1.1cm]{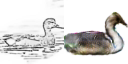} &
			\includegraphics[height=1.1cm]{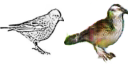} &
			\includegraphics[height=1.1cm]{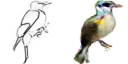} &
			\includegraphics[height=1.1cm]{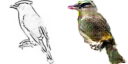} \\
			Sim &
			\includegraphics[height=1.1cm]{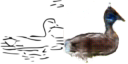} &
			\includegraphics[height=1.1cm]{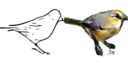} &
			\includegraphics[height=1.1cm]{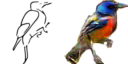} &
			\includegraphics[height=1.1cm]{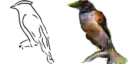} \\
		\end{tabular}
	\end{center}
	\caption{Results on CUB dataset in three sketch styles: XDoG, Photocopy (PC) and simplified (Sim). Best viewed in color.}
	\label{fig:bird_res}
\end{figure}

We further validate the robustness of the proposed framework using the CUB bird dataset. Compared to face and car, the CUB bird images contain much more texture information (e.g., the feather) which makes learning sketches as well as the correspondence more difficult. To get rid of the negative effect, we adopt relative total variation smoothing~\cite{tsmoothing2012} to preprocess the sketches which was designed to separate structure from texture. Then, we combine them with the original images to form our joint images. Results are shown in Figure~\ref{fig:bird_res}. 

\subsubsection{Car}
\begin{figure}[!t]
	\begin{center}
		\footnotesize
		\includegraphics[height=1.1cm]{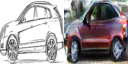} 
		\includegraphics[height=1.1cm]{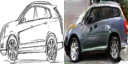} 
		\includegraphics[height=1.1cm]{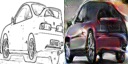} 
		\includegraphics[height=1.1cm]{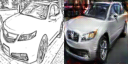} 
		\includegraphics[height=1.1cm]{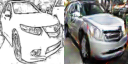}
		\includegraphics[height=1.1cm]{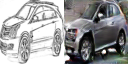}
		\includegraphics[height=1.1cm]{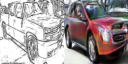}
		\includegraphics[height=1.1cm]{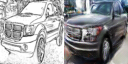}
		\includegraphics[height=1.1cm]{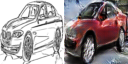}
		\includegraphics[height=1.1cm]{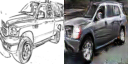} \\
	\end{center}
	\caption{Results on Car dataset in FDoG style.}
	\label{fig:car_res}
\end{figure}

To demonstrate that our framework is generic and can be applied to other categories of images, we also tested it on car images. Unlike face and bird, the car dataset is even more challenging because of the cluttered background and different poses of car. Figure~\ref{fig:car_res} shows sampled results using our approach. Note that the first two input sketches are identical. However, our network is able to generate two different images, i.e., cars with different shapes and colors while the input sketch still constrains the generated car features but not requiring them to be in strict alignment with the sketch. 

\begin{figure*}[t]
	\begin{center}
		\includegraphics[width=0.98\linewidth]{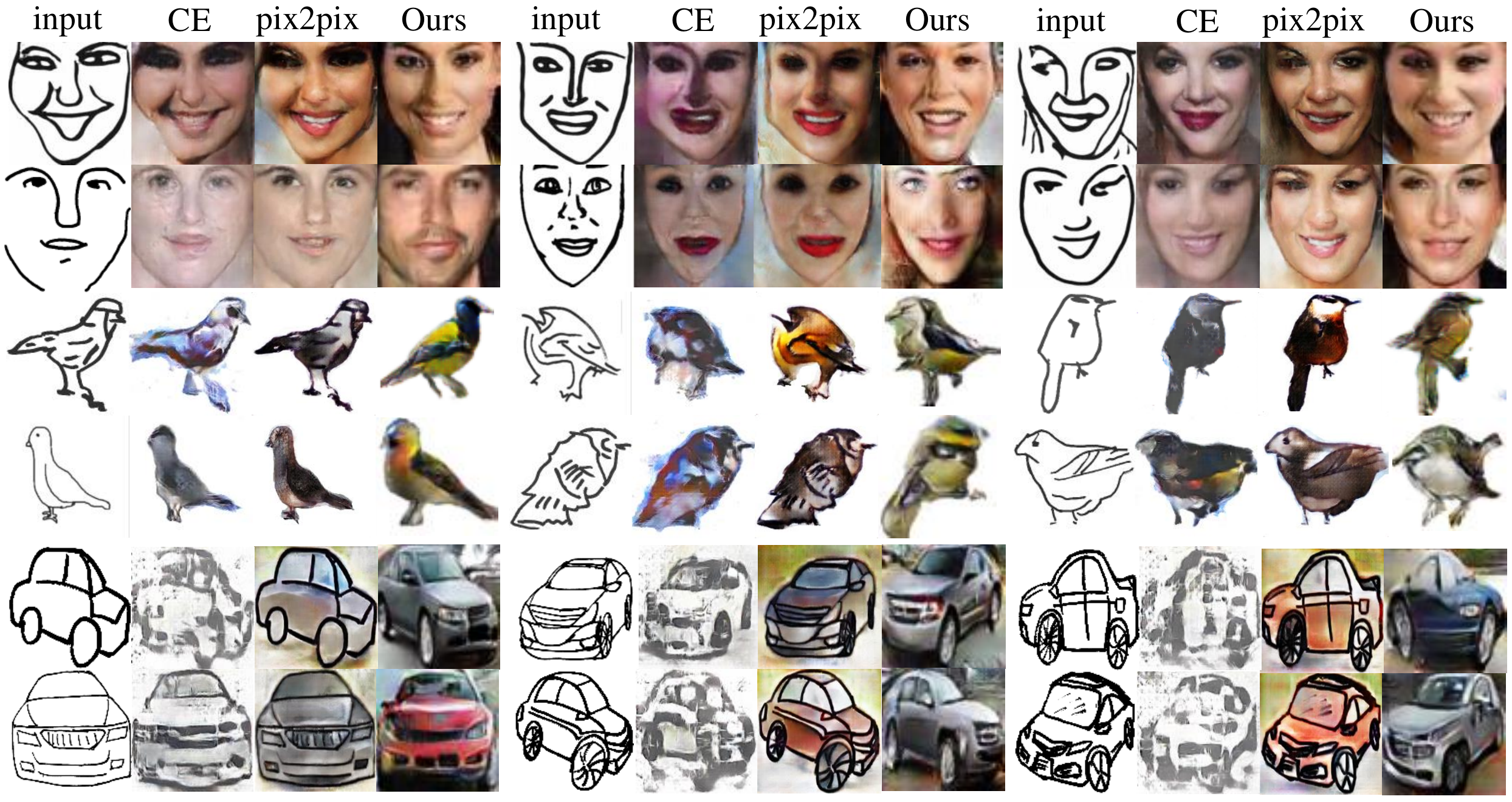} \\
	\end{center}
	\caption{Comparison with CE~\cite{contextencoder} and pix2pix~\cite{isola2016image} using 
		ugly/poorly drawn sketches on three different classes. For each group, from left to right: 
		input sketches, CE results, pix2pix results and our results. 
		Our method is resilient to corrupted/abstract inputs given in bad quality.
		Best viewed in color. 
	}
	\label{fig:pix2pix}
\end{figure*}

\subsubsection{Comparison on Hand-Drawn Sketches}
We also investigated our model's capability of generating images from (ugly) free-hand sketches. We collect 50 free-hand sketches for each of the 3 categories. Each of the sketches is drawn given a random image. We compare our results with Context Encoder (CE)~\cite{contextencoder} and Image-to-Image Translation (pix2pix)~\cite{isola2016image}. Figure~\ref{fig:pix2pix} demonstrates sample results. For fair of comparison, all methods are tested with the same style model (simplification for the face and bird datasets, FDoG for the car dataset) using the default parameter settings, without finetuning on the hand-drawn data. 

From Figure~\ref{fig:pix2pix} we learn that when directly applied to free-hand sketches without finetuning, pix2pix is unable to generate photo-realistic natural images given the ugly/abstract sketch inputs. Though it can learn accurate semantic contents, e.g., eyes, nose, mouth etc., it tends to strictly follow the shape of the input sketch even if the output falls far apart from the learned data distribution and has a high adversarial loss (in our case, the perceptual loss). CE is even worse when it comes to the car dataset. By contrast, our results show that the proposed framework is resilient to corrupted inputs given in bad quality, where we manage to map the input to the closest $z$ in the latent space and use this closest vector to generate images that reflect the semantics of the input sketch while looking natural and realistic. Note that pix2pix produces deterministic outputs with little stochasticity, while our method is able to produce stochastic output by updating the manifold $z$, which is likely to capture a fuller spectrum of the data distribution.
Figure~\ref{fig:manifold} gives an evidence by providing visualization of traversing the latent space during back-propagation (detailed in section~\ref{sec:two-stage}).

\subsection{Quantitative Evaluation}
While we clearly outperform CE~\cite{contextencoder} and pix2pix~\cite{isola2016image} 5on badly drawn sketches, 
we further conducted two quantitative experiments on good sketches for sake of fairness, where the edges correspond quite well to the corresponding photographic objects: (a) SSIM with ground truth; (b) face verification accuracy. Both (a) and (b) are evaluated on CelebA with 1000 test images.  

\begin{table}[t] 
	\centering
	\caption{SSIM and verification accuracy on CelebA test sets.}
	\begin{tabular}{c|c|c|c||c|c|c|c}
		\hline
		Method & pix2pix~\cite{isola2016image} & CE~\cite{contextencoder} & Ours & Method & pix2pix~\cite{isola2016image} & CE~\cite{contextencoder} & Ours \\
		\hline
		SSIM & 0.9012 & 0.5477 & \textbf{0.8856} & Verification Acc. & 99.69 & 97.19 & \textbf{99.80} \\
		\hline
	\end{tabular}
	\label{tab:ssim}
\end{table}

\noindent \textbf{SSIM}: The Structural similarity metric (SSIM)~\cite{wang2004image} is used to measure the similarity between generated image and ground truth. Results are shown in Table~\ref{tab:ssim}. We achieved comparable results to pix2pix on typical sketches, much better than CE. Note that outputs of pix2pix and CE strictly follow the input sketches and SSIM may fail to incorporate measures of human perception if input sketch is badly drawn.

\noindent \textbf{Verification Accuracy}: The motivation of this study is that if the generated faces are plausible it should have the same identity label with ground truth. The identity preserving features were extracted using the pretrained Light CNN~\cite{wu2015light} and compared using L2 norm. Table~\ref{tab:ssim} tabulates the results: we outperformed pix2pix, which shows that our model not only 
learns to capture the important details but is more resilient to different sketches. 

\subsection{Bi-directional Generation}
We also provide comparisons on forward generation, i.e., synthesizing a sketch from an image by corrupting the sketch portion. As we learn a joint distribution of the sketch and image, there is no difference whether we generate image from sketch, or sketch from image under our contextual GAN framework. 

\begin{figure}[t]
	\begin{center}
		\footnotesize
			\includegraphics[height=0.9cm]{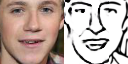} 
			\includegraphics[height=0.9cm]{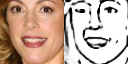} 
			\includegraphics[height=0.9cm]{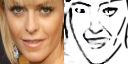} 
			\includegraphics[height=0.9cm]{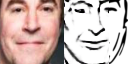} \\
			\includegraphics[height=0.9cm]{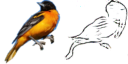} 
			\includegraphics[height=0.9cm]{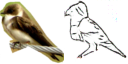} 
			\includegraphics[height=0.9cm]{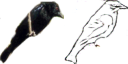} 
			\includegraphics[height=0.9cm]{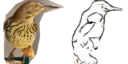} \\
			\includegraphics[height=0.9cm]{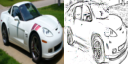} 
			\includegraphics[height=0.9cm]{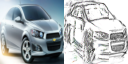} 
			\includegraphics[height=0.9cm]{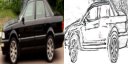} 
			\includegraphics[height=0.9cm]{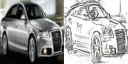} \\		
	\end{center}
	\caption{Reverse generation. Odd columns are input images, while even columns are generated sketches.}
	\label{fig:img2sketch}
\end{figure}

We adopt the same network architecture (see Sec.~\ref{sec:network}) and implementation (see Sec.~\ref{sec:implementation}) as in sketch-to-image scenario, except that we swap the role of sketch and image as the held-out portion in training and testing. Figure~\ref{fig:img2sketch} shows some convincing results on generating a sketch from 
a photographic image, which also demonstrates that our model can learn faithful
correspondence between a sketch and its corresponding image.

\subsection{Limitations}
\begin{figure}[t]
	\begin{center}
		\footnotesize
		\begin{tabular}{cccccc}
			\includegraphics[height=0.9cm]{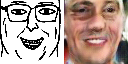} &
			\includegraphics[height=0.9cm]{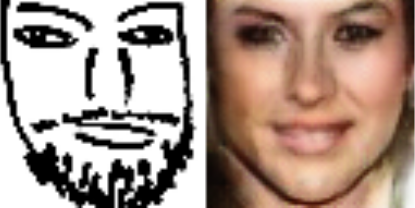} &
			\includegraphics[height=0.9cm]{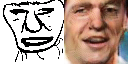} \\
			(a) & (b) & (c)
		\end{tabular}
	\end{center}
	\caption{Failure cases. (a) (b) missing attributes; (c) irregular shape of input face.}
	\label{fig:failure}
\end{figure}

Though our contextual GAN is resilient to ugly/abstract sketches and can realistically generate images which exhibit more freedom in appearance, one potential limitation is that in terms of face, we hope the generated image can preserve the identity of the input sketch (i.e., they represent the same person). However, due to the nature of freehand sketch, there is no guarantee of identity-preserving face generation given the sparse visual content. Also, it may fail to identify some kinds of attributes associated with the input. Figure~\ref{fig:failure}(a) and Figure~\ref{fig:failure}(b) visualize two cases of missing eye glasses and beard for the outputs, while they perceptually correspond to their inputs overall. We believe adding constraints like face attributes will better guide the generation process. We mainly focus on our proposed framework and leave it for future work. Figure~\ref{fig:failure}(c) shows another failure example with irregular shape of handdrawn face, which lies outside the data subspace, making it hard to find the closest mapping in the latent space.

\section{Conclusion and Future Work}
We show that the problem of sketch-to-image generation can be formulated as the {\em joint image completion} problem with the sketch providing the context for completion. Based on this novel idea we propose a new and simple contextual GAN framework. A generative adversarial network is trained to learn the joint distribution and capture the inherent correspondence between a sketch and its corresponding image, thus bypassing the cross-domain learning issues. This approach encodes the ``corrupted" joint image into the closest ``uncorrupted" joint image in the latent space, which can be used to predict and hence generate the output image part of the joint image. 

Compared with the end-to-end methods, our approach is a two-stage method that requires longer inferring time during testing. However, the two-stage approach allows us to separate the training and testing. In training, our generator learns the natural appearance of faces, cars, and birds such that any noise vector in the latent space would be able to generate a visual plausible image. On testing, although we have augmented the sketch drawing by three different sketch styles, we do not restrict the human free hand drawing to strictly follow the three augmented styles. 
We conduct thorough experiments to demonstrate the advantages of the proposed framework. 
In the future we plan to investigate more powerful generative models and explore more application scenarios. 
While our output is faithful to the input sketch, new quantitative measurement may be 
needed to measure the ``perceptual" correspondence between a (badly drawn) 
input sketch and our generated image (e.g.~Figure~\ref{fig:teaser}), a subject and difficult 
problem in its own right.

\subsubsection{Acknowledgement}
This work was supported in part by Tencent Youtu.

%
%
%

\bibliographystyle{splncs04}
\bibliography{sketchbib}

\end{document}